# HUNSPELL for Sorani Kurdish Spell Checking and Morphological Analysis

**Sina Ahmadi**
Insight Centre for Data Analytics
National University of Ireland Galway
ahmadi.sina@outlook.com

## Abstract

Spell checking and morphological analysis are two fundamental tasks in text and natural language processing and are addressed in the early stages of the development of language technology. Despite the previous efforts, there is no progress in open-source to create such tools for Sorani Kurdish, also known as Central Kurdish, as a less-resourced language. In this paper, we present our efforts in annotating a lexicon with morphosyntactic tags and also, extracting morphological rules of Sorani Kurdish to build a morphological analyzer, a stemmer and a spell-checking system using HUNSPELL. This implementation can be used for further developments in the field by researchers and also, be integrated into text editors under a publicly available license.[1]

## 1 Introduction

Spelling error detection and correction are respectively the two tasks of detecting a spelling error and correcting it by suggesting possible alternatives. These two tasks are generally carried out by a spell checking system which is built at early stages of development of language technology tools for a given language. Moreover, the analysis of the morphological structure of words is of importance as a subjacent component of natural language processing (NLP) pipelines where the morphemes of a given word are analyzed following morphological segmentation and tagging of composing morphemes. This task is beneficial for various purposes, particularly part-of-speech tagging, lemmatization and syntactic analysis.

Despite the recent advances in applying language-independent statistical and neural network methods, such as $n$-gram and sequence-to-sequence language modeling (Moeller et al., 2018; Ahmadi, 2018; Beemer et al., 2020), rule-based methods are still highly in use thanks to accuracy and viability, particularly in the case of morphologically rich languages like Kurdish. Among the rule-based methods, finite state transducers have been widely used for this task, as in (Fransen, 2020) for Old Irish, (Creutz and Lagus, 2002) for Finnish and (Nicholson et al., 2012) for Inuktitut. Although the tasks of morphological analysis, stemming and spell checking are addressed in previous studies for Kurdish, there is no open-source tool available (Ahmadi, 2020b). Therefore, in this paper, we present a rule-based approach to create a Sorani Kurdish morphological analyzer and spell checking system using HUNSPELL. HUNSPELL (Ooms, 2017) is a spell checker and morphological analyzer originally designed for languages with rich morphology and complex word compounding. An an open-source software, it is widely integrated by various web browsers and text editors for spell checking.

We previously described the Sorani Kurdish morphology from a computational point of view where the word-forms of lexemes with an open-class part-of-speech are defined (Ahmadi, 2021). These morphological rules are considered as finite-state transducers which can be used to analyze and generate various word-forms for a given lexeme. Using these rules, we define 4,293 morphological rules and demonstrate that the extracted rules can efficiently detect and suggest spelling errors and also, morphologically analyze words by stripping affixes, retrieving stems and extracting morphemes. Furthermore, we create a morphosyntactically-tagged lexicon containing 23,223 entries in Sorani Kurdish. The tagged lexicon along with the morphological rules are then used in HUNSPELL.

[1] https://github.com/sinaahmadi/KurdishHunspell

The rest of the paper is organized as follows. We first briefly discuss the previous efforts in the same field in Section 2. In Section 3, the essential features of Sorani Kurdish morphology are described. In Section 4, we describe the development of the lexicon and the implementation of an affix file in Hunspell based on morphological rules to detect word forms. We present our experiments on spell-checking, stemming and morphological analysis in Section 5. Finally, the paper concludes in Section 6 where the main contributions of the current work along with a few suggestions for the future work are highlighted.

Throughout this manuscript, we use the Leipzig glossing rules (Comrie et al., 2008) for interlinear morpheme-by-morpheme glosses. We also mark the clitic and affix boundaries respectively with '=' and '-'. Kurdish words are also provided in both Arabic-based and Latin-based scripts.

## 2 Related Work

### 2.1 Morphological Analysis

Some aspects of Kurdish morphology and syntax have been previously studied by linguists. Excluding pedagogical resources describing general grammar of Sorani Kurdish, we can mention the following major contributions to the field. To further clarify, an overview of Kurdish linguistics is provided by Haig and Matras (2002).

Among the analytic resources describing Sorani Kurdish morphology, we can mention (Hacî Marif, 2000; Blau, 2000; McCarus, 1958) where various morphological and syntactical features of Sorani Kurdish dialects are studied. On the interaction of morphology and phonology, Abdulla (1980) and Rashid Ahmad (1987) analyze some aspects of the Kurdish phonology where various morphological rules are taken into account. One of the specific characteristics of Kurdish is ergatavity. This topic has been also widely studied in Indo-Iranian languages in general (Karimi, 2012), and Kurdish in particular (Jukil, 2015; Mahalingappa, 2013; Karimi, 2010; Karimi, 2014). Another interesting topic is the *Ezafe* (or *Izafa*) construction which has been addressed as well (Karimi, 2007; Sahin, 2018).

Focused on specific morphemes, Baban (2001) studies the repetition suffix *-ewe* in Sorani Kurdish and analyzes both morphology and syntax with respect to this morpheme. In the same vein, Doostan and Daneshpazhouh (2019) focus on the *-râ* morpheme from syntactic and semantic perspectives. Another related work to the current project is (McCarus, 2007) which analyzes the inflectional and derivational patterns in the Sorani Kurdish morphology. Fatah and Hamawand (2014) provide descriptions of some of the Sorani Kurdish positive and negative prefixes based on the Prototype Theory. Karacan and Khalid (2016) analyze adjectives in various Kurdish dialects. Walther (2012) investigates morphological structures of Sorani Kurdish mobile person markers. The work extends the formal analysis of Sorani Kurdish endoclitic placement by Samvelian (2007).

Although the knowledge of morphology for creating linguistic resources is of fundamental importance in NLP and computational linguistics, there are only a limited number of studies that deal with Kurdish computational morphology, directly or explicitly within other tasks. As the earliest attempt in generative grammar for Kurdish, Baban and Husein (1995) provide a description to create Kurdish sentences by focusing on both morphology and syntax. Despite its originality and novelty, this study does not seem to be widely used or followed by the succeeding studies. As such, Walther and Sagot (2010) and Walther et al. (2010) present a methodology and preliminary experiments on constructing a morphological lexicon in the Alexina Framework (Sagot, 2010) for both Sorani and Kurmanji in which a lemma and a morphosyntactic tag are associated with each known form of the word. The lexicon[2] contains 22,990 word forms of 274 lemmata and is added to the Universal Morphology (UniMorph) project[3] which is a collaborative effort to improve how NLP handles complex morphology in the world's languages (Kirov et al., 2018; Cotterell et al., 2017). One limitation regarding this study is the lack of coverage of complex morphological forms such as inflected verbs based on tenses and aspects. Moreover, we noticed that many of the word forms are not correct and some follow the morphology of a specific sub-dialect of Sorani Kurdish and are not globally used in the dialect. More recently, Ahmadi and Hassani (2020) present a preliminary study on creating a morphological analyzer using finite-state transducers. Furthermore, (Ahmadi, 2020a) creates a morphological analyzer for the task of lexical analysis, also known as tokenization, for both Sorani and Kurmanji dialects of Kurdish where the extraction of morphemes is taken into account as a sub-task.

[2] https://github.com/unimorph/ckb
[3] https://unimorph.github.io

### 2.2 Spelling Error Correction and Stemming

Similarly, this task has been previously addressed in a few studies. Salavati and Ahmadi (2018) present a system for lemmatization and spelling error correction using $n$-gram models and a non-exhaustive list of the most frequent morphemes in Sorani Kurdish. Similarly, Hawezi et al. (2019) propose an algorithm for Sorani Kurdish spell checking using a lexicon and a set of patterns to define word forms. A given entry is then corrected by calculating the Levenshtein distance. Saeed et al. (2018a) and Mustafa and Rashid (2018) propose a stemming system for Sorani Kurdish which retrieves word stems by removing longest suffix and prefixes of words. Salavati et al. (2013) present a rule-based and statistical stemmer for both Sorani and Kurmanji Kurdish. Saeed et al. (2018b) utilizes the Porter stemming algorithm for Kurdish text classification. The previous studies on stemming are carried out in the context of information retrieval and are reportedly efficient in stemming queries. More importantly, Amani and Koochari (2017) and Mahmudi (2019) create spell checkers for Sorani, respectively, using the Soundex algorithm and a statistical approach, and report their efficiency when evaluated on their test sets. The current study is compared with these two spell checking approaches.

Based on our literature review, no morphological analyzer or spelling error detection and correction system is openly available for Sorani Kurdish.[4]

## 3 Sorani Kurdish Morphology

Bound morphemes:
- Clitics
  - Enclitics : *=im*, *=î*, *=ît*, *=e*, *=ê*, *=êt*, *=în*, *=in*, *=wa*, *=wat*, *=a*, *=at*, *=te*
  - Endoclitics: *=im*, *=it*, *=î*, *=man*, *=tan*, *=yan*, *=îş*
- Affixes
  - Inflectional
    - Noun:*-êk*, *-an*, *-gel*, *-ha*, *-at*, *-eke*, *-ekan*, *-e*, *-ane*, *-gele*, *-î*, *-ê*, *-îne*
    - Verb:*-in*, *de-*, *e-*, *bi-*, *we-*, *ne-*, *na-*, *me-*
    - Adjective:*-tir*, *-tirîn*
    - Adverb:*-ane*
  - Derivational:*-a*, *-and*, *-ard*, *-d*, *-îşt*, *-ird*, *-t*, *-û*, *-y*, *-ên*, *-êr*, *-diran*, *-ran*, *-ewe*, *ber*, *da*, *der*, *heł*, *ser*, *řa*, *řo*, *wer*, -∅

Figure 1: A classification of Sorani Kurdish bound morphemes with a few instances

Kurdish is an Indo-European language spoken by over 25 million speakers in the Kurdish regions in Turkey, Iran, Iraq and Syria, and also by the Kurdish diaspora around the world (McCarus, 2007). Sorani Kurdish, also known as Central Kurdish (`ckb` in ISO 639-2), is the Kurdish dialect which is mostly spoken by the Kurds within the Iranian and Iraqi regions of Kurdistan. Sorani Kurdish is a null-subject language and has a subject-object-verb (S-O-V) order and can be distinguished from other Indo-Iranian languages by its ergative-absolutive alignment which appears in past tenses of transitive verbs (Ahmadi and Masoud, 2020). In this section, we provide a brief description of the Sorani Kurdish morphology by focusing on morphemes. Morphemes are classified into free and bound. While free morphemes are meaningful as they are, bound morphemes only carry meaning when affixed with other words.

[4] As of September 2021

### 3.1 Bound Morphemes

Bound morphemes are classified into two categories of affixes and clitics. Affixes and clitics are similar in the way that they cannot constitute a word and they lean on a prosodic host, i.e. a word for stress assignment. Clitics can appear with hosts of various syntactic categories while affixes only combine with syntactically-related stems (Haspelmath and Sims, 2013). The clitics and affixes in Sorani Kurdish have been widely studied previously and have been shown to be challenging considering the general theory of clitics (W. Smith, 2014; Gharib and Pye, 2018). This problem is particularly observed with respect to the direct and oblique person markers which can appear in different positions within a word-form depending on the functionality. In this section, the clitics and affixes in Sorani Kurdish are described. Figure 1 provides the most frequent clitics and affixes in Sorani Kurdish.

#### 3.1.1 Clitics

Clitics are categorized based on their position with respect to the host. A clitic is called proclitic and enclitic, if it appears before and after the host, respectively. There are two other forms of clitics which are non-peripherical and exist only among a few number of natural languages. If a clitic appears between the host and another affix, it is called a mesoclitic. A different type of non-peripheral clitic is endoclitic which appears within the host itself and is unique to a few languages around the world, such as Udi (W. Smith, 2014), Degema (Kari, 2002) and also Sorani Kurdish.

Sorani Kurdish has two types of endoclitics: pronominal makers (also introduced as mobile person markers by (Walther, 2012)) and the emphasis endoclitic یش =*îş* which can be translated as 'also' or 'too'. The pronominal endoclitics function as agent markers for transitive verbs in the past tenses or endoclitics as patient marker for transitive verbs in the present tenses. This is due to the split ergativity feature of Sorani Kurdish where the agent and patient markers are specified differently. The following examples show the alignment in present and past tenses of کەوتن (KEWTIN, 'to fall') and گرتن (GIRTIN, 'to get'). The agent marker ن *-in* in intransitive present tenses serve as patient marker in transitive past tenses due to ergativity.

(1) *deke**win*** دەکەون
*de-kew-**in***
fall.PRS.PROG.INTR.3PL
'(they) are falling.'

(2) *degi**rin*** دەگرن
*de-gir-**in***
get.PRS.PROG.TR.3PL
'(they) are getting.'

(3) *kewt**in*** کەوتن
*kewt-**in***
fall.PST.PROG.INTR.3PL
'(they) fell.'

(4) *girtyan**in*** گرتیانن
*girt=yan-**in***
get.PST.PROG.TR.ERG.3PL.3PL
'(they) got (them).'

| | | | | | | | | | | | | |
|---|---|---|---|---|---|---|---|---|---|---|---|---|
| 0 | | | | | | girt | | | | | | past stem of GIRTIN (to take, to get) |
| 1 | | | | | | girt | im | | | | | I got. |
| 2 | | | | | | girt | im | in | | | | I got them. |
| 3 | | | | | | girt | im | in | e | | | I got them to/with. |
| 4 | | | | | | girt | im | in | e | ewe | | I got them to/with again. |
| 5 | | | | | | girt | îş | im | in | e | ewe | I got them also to/with again. |
| 6 | | | ne | îş | im | girt | in | e | ewe | | | I did not get them also to/with again. |
| 7 | | ne | îş | im | de | girt | in | e | ewe | | | I was not getting them also to/with again. |
| 8 | da | îş | im | ne | de | girt | in | e | ewe | | | I was not taking down them also to/with again. |

Table 1: The placement of the endoclitic =*îş* (in green boxes) and agent marker =*im* (in blue boxes) with respect to the base and each other in a verb form. Note that Sorani Kurdish is a null-subject language.

Furthermore, the two endoclitic categories of Sorani appear in an erratic pattern within a word form or a phrase. Table 1 presents an example where the 1SG marker م (=*im*) and the emphasis endoclitic =*îş* appear after and before the host, i.e. گرت (*girt*), depending on the presence of other bound morphemes such as negation prefix نە (*ne-*) or the verbal particle دا (*da*). It is worth mentioning that this pattern may vary based on the Sorani sub-dialects.

### 3.1.2 Affixes

In comparison to clitics, a bigger number of bound morphemes in Sorani Kurdish belong to affixes. Affixes can be categorized into inflectional and derivational based on their ability to create new lexemes. The most frequent affixes in Sorani appear as prefix and suffix. Some of the inflectional affixes of Sorani Kurdish belonging to open-class parts of speech, namely nouns, verbs, adjectives and adverbs, are shown in Table 2. In addition, Izafa particle *-î* and its allomorph *-e* which appear between a head and its dependents in a noun phrase are frequently used to create possessive constructions, as in ناوی من (*nêw-î min* 'my name').

| Nouns | Verbs | Adjectives | Adverbs |
|---|---|---|---|
| **number** (SG, PL)<br>**person** (1, 2, 3)<br>**determiners** (DEF, IND, DEM)<br>**case** (OBL, LOC, VOC)<br>**gender** (M, F) | **number** (SG, PL)<br>**person** (1, 2, 3)<br>**mood** (IND, SBJV, IMP, COND)<br>**aspect** (PRF, IMP, PROG)<br>**tense** (PST, PRS) | **number** (SG, PL)<br>**degree** (COMP, SUPL)<br>**determiners** (DEF, IND, DEM) | **degree** (COMP, SUPL) |

Table 2: Inflectional features and values of Sorani Kurdish. It should be noted that the function of cases and genders vary among Sorani subdialects.

In addition to compound words, Sorani Kurdish relies on derivational morphemes to create new lexemes, particularly new verbal lexemes. To this end, verbal suffix ەوە (*-ewe*) and verbal particles such as دا (*da-*) and هەڵ (*heł-*) are used. It is worth noting that passive form of verbs is derived from the verb stem by using دران/درێ (*-diran/-dirê*) (or their allomorphs ران/رێ (*-ran/-rê*) (Doostan and Daneshpazhouh, 2019)) suffixes, unlike Kurmanji Kurdish which relies on periphrastic forms with HATIN (to come).

A more detailed description of Sorani Kurdish morphology, including adpositions and pronouns as free morphemes is provided in (Ahmadi, 2021).

## 4 Implementation

As a rule-based method, HUNSPELL needs an annotated lexicon to which the morphological rules are applied. The lexicon (with `.dic` extension) contains words with morphosyntactic flags, and the morphological rules are included in the affix file (with `.aff` extension) that specifies how word forms are created based on the word flags in the lexicon. The creation of the lexicon and implementation of the rules are described in this section and a few samples of these two files are provided in Appendix A.

### 4.1 Lexicon Annotation

Since there is no annotated lexicon available for this purpose, we use the Sorani Kurdish lexicographic material provided by Wîkîferheng, the Kurdish Wiktionary[5] and the FreeDict project[6]. The latter provides the content in the Latin-based script of Kurdish. Therefore, we use the rule-based transliteration system provided by (Ahmadi, 2019) to transliterate it into the Arabic-based script which is used in our implementation. In addition, Wikidata is consulted to extract proper names using the query provided in Appendix B. Each lemma in the lexicon is manually tagged with information such as part-of-speech, its formation type, i.e. derivational or inflectional, and further morphological properties such as stem, person and tense. In addition, composing parts of compound forms are specified using a hyphen which can be beneficial for lexical analysis. The number of entries per morphosyntactic tags are provided in Table 3.

[5] `https://ku.wiktionary.org`
[6] `https://freedict.org`

### 4.2 Morphological Rules

The morphological rules in Hunspell are defined as prefixes and suffixes, respectively specified with `PFX` and `SFX`, in such a way that new forms of the lemmata and stems in the lexicon are analyzed or generated by stripping these morphemes from a given word-form or adding them to the lemmata or stems in the lexicon. In addition, it is possible to integrate the morphophonological features which create the surface realization of word forms based on the phonology of the language. Following the morphophonological rules described in (Ahmadi, 2021), automatic and morphophonological alternations are integrated in the morphological rules in the affix file. Although Hunspell allows specifying automatic alternations in brackets ([]) by defining different rules for a specific suffix if preceded by a vowel or consonant, as provided in Appendix A, morphophonological alternations are defined as separate rules and in some cases may require separate entries in the lexicon as well. For instance, the morpheme ڕەوە (-*rewe*) in بخوێنەرەوە (*bixwênerewe*, READ.V.IMP.2SG) is a morphophonological alternation of the morpheme ەوە (-*ewe*) which is therefore defined as a new rule.

Moreover, although Hunspell allows twofold prefix stripping, we only define single stripping rules for both prefixes and suffixes to facilitate the maintenance and future developments. In other words, the set of possible combinations of prefixes or suffixes that can appear with a base are first detected and then defined in Hunspell as individual rules. Table 4 summarizes the number of rules in our implementation based on their morphosyntactic categories. It should be noted that prefix and suffix in this context refer to the position of the morphemes with respect to the host. Therefore, clitics are also included in the rules similar to affixes.

In order to be able to use Hunspell for morphological analyzer, we use a wrapper program[7].

| Tag | Part-of-speech | # lemmata |
|---|---|---|
| `N` | noun | 20,538 |
| `V` | present stem of a verb | 1,005 |
| `I` | past stem of intransitive verb | 779 |
| `T` | past stem of transitive verb | 271 |
| `A` | adjective | 1,732 |
| `R` | adverb | 441 |
| `E` | numeral | 51 |
| `C` | conjunction | 38 |
| `D` | interjection | 82 |
| `B` | pronoun | 30 |
| `E` | numeral | 51 |
| `F` | adposition | 13 |
| `G` | particle | 16 |
| `X` | infinitive | 3,528 |
| `Z` | proper names | 9,893 |
| `W` | exceptional cases like *were* ’come.imp.2s’ | 39 |
| All | | 23,223 |

Table 3: Statistics of the annotated lexicon

| Tag | Part-of-speech | Prefix (`PFX`) | Suffix (`SFX`) | All |
|---|---|---|---|---|
| `N` | noun | - | 913 | 913 |
| `A` | adjective | - | 731 | 731 |
| `V` | present tenses | 875 | 30 | 905 |
| `P` | passive | - | 239 | 239 |
| `I` | past tenses of intransitive | 107 | 126 | 233 |
| `T` | past tenses of transitive | 830 | 414 | 1,244 |
| `E` | numeral | - | 7 | 7 |
| `B` | pronoun | - | 21 | 21 |
| All | | 1812 | 2481 | 4293 |

Table 4: Statistics of the morphological rules for Sorani Kurdish

## 5 Experiments

### 5.1 Spelling Error Correction

**Test Sets** Sorani Kurdish lacks a unified orthography. To tackle this problem, we follow the guidelines provided by (Hashemi, 2016) as our orthographic reference. These guidelines are similarly taken into account in the two test sets which are used by (Amani and Koochari, 2017) and (Mahmudi, 2019). Therefore, we use these test sets and respectively refer to them as $T_1$ which contains 385 instances among which there is no correctly spelled word, and $T_2$ which contains 1,400 instances among which %60.6 are correct. Figure 2 illustrates a description of this dataset. In addition, $T_2$ provides more challenging incorrectly spelled instances, such as those due to missed spaces, i.e. two or more separate words merged together without a space. Such cases are specified as "incorrect (spaced)" in Figure 2. We note that the spelling errors in $T_2$ contain more inflected forms such as plural forms, conjugated verbs and various verb tenses in comparison to $T_1$.

[7]`https://github.com/MSeal/cython_hunspell`

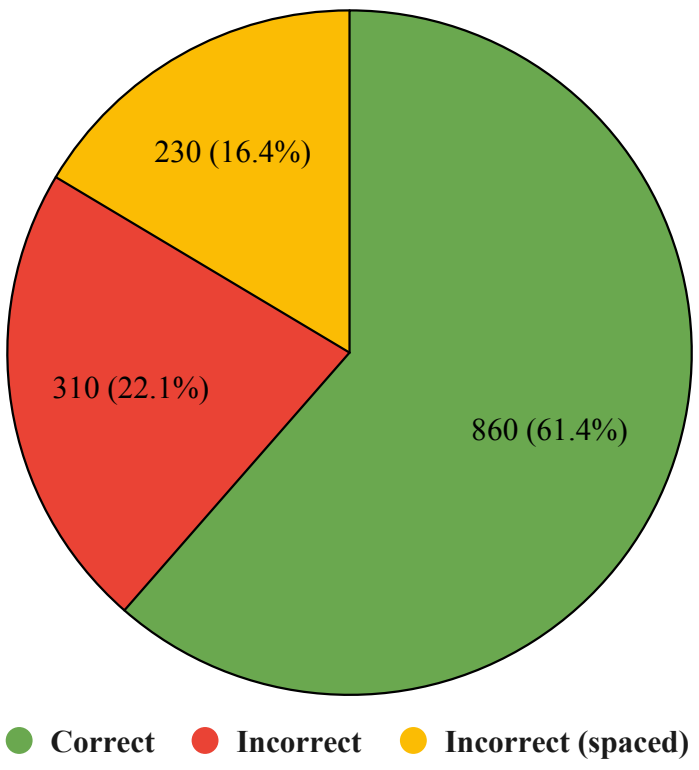

Figure 2: A description of Mahmudi (2019)'s dataset which is used for evaluating the HUNSPELL spell checking of Kurdish in this paper. Among the correctly spelled cases, no word is found with a space.

**Baseline System** Furthermore, we extract the 2,376,405 unique word forms from Sorani Kurdish corpora, namely those provided by (Esmaili et al., 2013; Veisi et al., 2020; Ahmadi et al., 2020) among which words with a frequency of 10 or more are considered as potentially correct words that can be used to create a baseline system. Such words make almost 11% of the whole extracted words, i.e. 265,216 words. In the baseline system, spelling errors are detected by looking up in the list of the words and also corrected by calculating the Levenshtein distance (Levenshtein, 1966) of the word with the unique words from the corpora. This way, if an input word appears in the list of the frequent words, it is considered correct, otherwise the baseline system provides suggestions sorted based on the edit distance.

**Evaluation** Table 5 provides the performance of the baseline system and our implementation in HUNSPELL along with the reported performance of the previous systems.[8] We calculate accuracy ($Acc$), Lexical Precision ($P_c$), Lexical Recall ($R_c$) and $F_1$ measures as defined by (Van Huyssteen et al., 2004) and based on the followings:

- true positive (TP) for correctly detected correct words.
- false positive (FP) for incorrectly detected correct words.
- true negative (TN) for correctly detected incorrect words.
- false negative (FN) for incorrectly detected incorrect words.

In addition, we calculate the existence of a correct prediction of an incorrect word based on their ranking, as in the first, three and more than three first suggestions of the systems. These cases are respectively denoted by $Sugg_1$, $Sugg_3$ and $Sugg_{>3}$. It should be noted that the correction suggestions are ranked by the most probable suggestions, i.e. the most probable replacement is supposed to be suggested first.

Our implementation in HUNSPELL performs differently and disparately with respect to each test set with an overall efficiency below the baseline system and that of the previous systems. Quantitatively, in all cases, our system fails to significantly detect correct words correctly leading to a lower precision, and also, fails to remarkably detect correct words among all the words, leading to a lower recall. On the other hand, the accuracy also indicates the competency of the spell checker in handling a given word accurately, i.e. the likelihood that the suggested words of the spell checker for a correct or incorrect input are correct. In the current version of our implementation, HUNSPELL is only accurate in 40% of the cases. Nevertheless, it is not clear to us whether the reported lexical precision and recall in the previous studies are calculated based on the prediction of correct instances or incorrect ones. Figure 3 shows a comparison of our system along with the baseline and the previous systems based on F1 score and the existence of a correct prediction in the suggestions, i.e. $Sugg_{>3}$.

[8] Please note that the evaluation of our system is carried out on Version 0.1.0. See `https://github.com/sinaahmadi/KurdishHunspell`

| System | Test set | TP | FP | TN | FN | Acc (%) | $P_c$ | $R_c$ | F1 | $Sugg_1$ (%) | $Sugg_3$ (%) | $Sugg_{>3}$ (%) |
|---|---|---|---|---|---|---|---|---|---|---|---|---|
| Baseline | $T_1$ | 0 | 0 | 178 | 206 | | | | | 26.82 | 37.76 | 46.35 |
| | $T_2$ | 597 | 251 | 105 | 441 | 50.36 | 0.70 | 0.58 | 0.63 | 10.87 | 15.76 | 19.02 |
| | $T_2\backslash$space | 597 | 251 | 105 | 210 | 60.36 | 0.70 | 0.74 | 0.72 | 18.69 | 27.10 | 32.71 |
| (Amani and Koochari, 2017)[9] | $T_1$ | | | | | | 0.98 | 0.97 | 0.98 | 68.17 | 73.94 | 74.94 |
| (Mahmudi, 2019)[10] | $T_1$ | 357 | 0 | 15 | 0 | 100 | 1.00 | 1.00 | 1.00 | 55.65 | 75.71 | 93.50 |
| | $T_2$ | 539 | 108 | 753 | 0 | 92.29 | 0.83 | 1.00 | 0.91 | 69.57 | 80.33 | 84.97 |
| Our system | $T_1$ | 0 | 0 | 208 | 176 | | | | | 22.92 | 39.84 | 47.66 |
| | $T_2$ | 367 | 481 | 182 | 364 | 39.38 | 0.43 | 0.50 | 0.46 | 18.30 | 26.99 | 31.52 |
| | $T_2\backslash$space | 367 | 481 | 93 | 222 | 39.55 | 0.43 | 0.62 | 0.51 | 16.51 | 24.30 | 27.73 |

Table 5: The performance of our implementation in HUNSPELL in comparison with the reported performance of the previous studies. The gray cells correspond to unreported scores in the related study or undefinable ones.

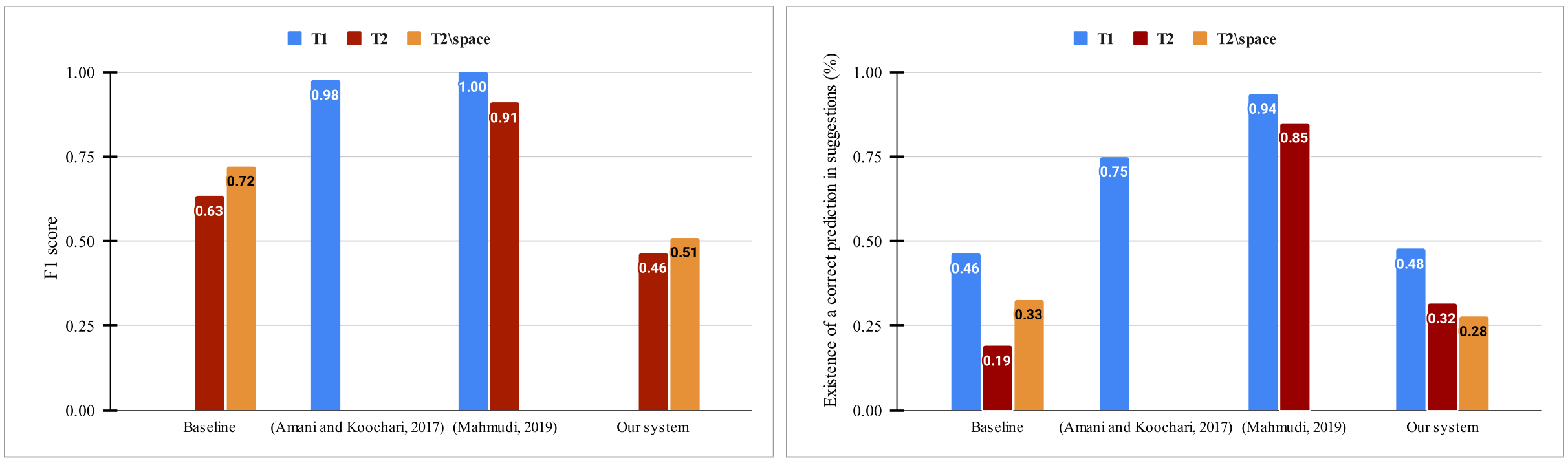

Figure 3: A comparison of the performance of our systems along with the two previous studies based on the F1-score and the existence of a correct prediction, i.e. $Sugg_{>3}$

From a qualitative point of view, the results of the evaluation of our system can be explained by the following issues:

(a) non-existence of a word of the test set in the annotated lexicon, e.g. هۆلیوود ‘Hollywood’. It should be noted that there are non-Kurdish words in the test sets, for instance words in Gorani or Arabic, which does not exist in our Sorani lexicon and therefore, cannot be correctly predicted.

(b) incorrect morphological synthesis of the input word or incorrect morphological generation of a suggested word, e.g. دەت واین (*det wayn*) as an incorrect suggestion for دەتواین (*detwayn*) instead of دەتوانین (*detwanîn*, ‘(we) can‘).

(c) lack of lexical variations in the test sets, e.g. تەنیا (*tenya*, ‘alone (adjective)’) is considered an incorrect spelling of تەنها (*tenha*) in (Mahmudi, 2019)’s test set, while both words are used in various subdialects of Sorani and thus, exist in our lexicon and are detected as correct.

(d) different convention regarding spacing as in دەریبکەن (*derîbiken*, ‘(that you.plural) dismiss’) which is a compound verb and can be also written with spaces in between composing parts as دەری بکەن (*derî biken*). However, such spacing is deemed incorrect in (Mahmudi, 2019)’s test set, while our system detects it as a correct spelling. Mahmudi (2019)’s test set is curated with the assumption that all compound forms in Kurdish are written contiguously without any spaces.

[9]The results of this system is based on the reported scores in (Mahmudi, 2019)
[10]See footnote 9

To further examine the last aforementioned point, we evaluate our baseline and HUNSPELL spell checkers on those words in $T_2$ for which spelling errors are not due to missing spaces, i.e. the red and green portions of Figure 2. This test set is referred to as $T_2\backslash$space in Table 5. Unlike the baseline system, the results indicate that HUNSPELL performs with the same precision of 0.43 but the higher recall of 0.62 when tested on $T_2\backslash$space in comparison to $T_2$. This can be explained by the lower number of incorrect words in $T_2\backslash$space which increases the likelihood of detecting an input word as correct (73.50% correct words in $T_2\backslash$space in comparison to 61.4% in $T_2$). Given the insignificance of difference between other scores, $Sugg_{>3}$ seems to be the only representative measure that indicates a lower percentage of correct suggestions (27.73%) by HUNSPELL on $T_2\backslash$space that is lower than $Sugg_{>3}$ of $T_2$ (31.52%).

Based on this qualitative analysis, we believe that our systems carry out the task of spell checking efficiently but differently from the expected corrections in the test sets, particularly in the case of compound words. It is worth mentioning that the Arabic-based scripts are known to present many challenges in spelling complexity (Habash et al., 2018). Ahmadi (2020a) tackles inconsistencies of spacing for Kurdish compound words differently for lexical analysis. Moreover, one major limitation of HUNSPELL is ignoring spaces in the input words. For instance, if the input word بەهار (*behar* 'spring (noun)') is mistakenly split into two tokens بە هار (*be har*), the spell checker cannot detect that the space causes an error. Given that there is no such type of error in the test sets, we believe that this feature of our implementation in HUNSPELL should be evaluated with a more representative test set.

Appendix C provides a few examples of the outputs of our baseline system and the implementation in HUNSPELL. To better describe the type of errors that the systems made, various types of incorrect words and predictions are specified in different colors.

### 5.2 Morphological Analysis

Although the Unimorph project provides a dataset for Sorani Kurdish morphological analysis[11], this dataset contains many incorrect word forms which seem to be generated without sufficient knowledge to assess Sorani Kurdish morphotactics. Therefore, we create a new test set as a preliminary effort to analyze Sorani Kurdish morphology. We annotate 140 word forms with information such as lemma, part-of-speech tag, stem, base, prefixes, suffixes and additional description. In this context, prefixes and suffixes respectively refer to the preceding and succeeding segments that appear before and after the base. Among the 140 word forms, 14 are adjectives, 16 are adverbs, 27 are nouns, 73 are verbs and the remaining 10 words have two or more part-of-speech tags.

We evaluate our morphological analyzer in HUNSPELL from the following three perspectives:

- Morphological segmentation which breaks a word into its composing morphemes (Poon et al., 2009). Evaluation results indicate that HUNSPELL has an accuracy of 80.14% to retrieve the base of a given word along with its preceding and succeeding morphemes.
- Analysis coverage: based on the unique word forms extracted from the corpora (discussed in Section 5.1), we calculate the number of word forms that can be segmented by our morphological analyzer, correctly or incorrectly. This indicates the extent to which morphological structures are known to the morphological analyzer and therefore, it can reveal the ability of the system to deal with the complexity of word forms. To this end, the extracted forms from the corpora are tokenized and preprocessed using KLPT (Ahmadi, 2020b). Among the 235,210 unique word forms, the morphological analyzer can analyze 27.16% of the words. We believe that this is due to a lack of lexical entries in our lexicon, and also many morphological patterns particularly from specific sub-dialects of Sorani Kurdish. Nevertheless, not all the extracted word forms are necessarily correctly spelled or valid.
- Part-of-speech tagging and stemming: Our implementation in HUNSPELL assigns a correct part-of-speech tag with an accuracy of 86.02%. Likewise, the stemming of verbs is carried out with an accuracy of 63.75%.[12]

[11] Available at `https://github.com/unimorph/ckb`

[12] Only stemming of verbs is possible in the current version.

We believe that the lower performance of the morphological analysis is due to undefined morphological rules of more challenging word forms such as نەشیانبینیبووم (*neşyanbînîbûm*, ‘(they) even had not seen me’), where the endoclitic =*ş* appears before the negation prefix *ne*- and the personal marker =*yan* (bîn is the host in this word). In addition, some word forms are more ambiguous to be analyzed. For instance, تێکۆشانم (*têkoşanim*, ‘my efforts’) is incorrectly tagged as a verb by Hunspell due to the host being the infinitive تێکۆشان (*têkoşan*, ‘to try’). Likewise, compound words are error-prone due to the space, as in پێیدا (*pêyda*) which is an absolute adposition but is analyzed similar to پێی دا (*pêy da*, ‘he/she gave it (to) him/her’). It is worth mentioning that the lemmas of the word forms in our test set exist in the lexicon, making the inaccuracy due to unseen words less significant.

## 6 Conclusion and Future Work

In this paper, a morphological analyzer and spell checker are introduced for Sorani Kurdish using Hunspell. Although morphological analysis and spell checking have been previously addressed for Sorani Kurdish, the literature review indicates that no tool is publicly available. As such, the current study describes tremendous efforts for defining morphological rules and tagging a lexicon to create an efficient and open-source tool for Sorani Kurdish. We demonstrate that our implementation performs efficiently for morphological analysis, spelling error detection and correction and also stemming. Given the complexity of Sorani Kurdish morphology, we believe that our tool and resources are useful to pave the way for further developments in Kurdish language processing. We also hope that our tool raises awareness among Kurdish users regarding writing and ultimately, help the tool to facilitate the standardization of the Sorani Kurdish orthography.

One major limitation in the tasks of morphological analysis and spelling error correction of Kurdish is the lack of benchmarking. This should be addressed by creating evaluation datasets that distinguishes between sub-dialects of Sorani, particularly Northern and Southern Sorani, and provides a finer representation of different spelling errors, especially for words in context rather than tokens. Therefore, as future work, the creation of a gold-standard dataset for evaluating spell checking systems for Kurdish is required. Moreover, addressing the morphology of subdialects of Sorani Kurdish and also, other variants of Kurdish to create similar tools and resources is necessary. We believe that our test set for evaluating Sorani Kurdish morphology should be further enriched with more complex word forms in a comprehensive way. We suggest that more derivational and inflectional morphemes to be added to the current implementation and also, the lexicon be further enriched over time. Based on the current study, more advanced NLP tasks, such as syntactic analysis, can be addressed in the future.

It is worth noting that both the spell checker and the morphological analyzer are openly available and operaional in the Kurdish Language Processing Toolkit.[13]

## 7 Acknowledgements

The author would like to greatly thank those who supported open-source projects in Kurdish language technology through GitHub Sponsors.[14]

[13] `https://github.com/sinaahmadi/klpt`

[14] `https://github.com/sponsors/sinaahmadi`

# Appendices

## A Implementation

| | |
|---|---|
| ئاخێن/V **po**:verb **is**:present_stem_transitive_active **st**:ئاخ | SFX N 0 [^ایێە] ەکەم |
| ئاخێندر/V **po**:verb **is**:present_stem_intransitive_passive **st**:ئاخ | SFX N 0 [^ایێە] ەکەت |
| ئاخێندرا/I **po**:verb **is**:past_stem_intransitive_passive **st**:ئاخ | SFX N 0 [^ایێە] ەکەی |
| ئاخێندران/XN **po**:verb **is**:infinitive_intransitive_passive **st**:ئاخ | SFX N 0 [^ایێە] ەکەمان |
| ئاخێنر/V **po**:verb **is**:present_stem_intransitive_passive **st**:ئاخ | SFX N 0 [اە] کەم |
| ئاخێنرا/I **po**:verb **is**:past_stem_intransitive_passive **st**:ئاخ | SFX N 0 [اە] کەت |
| ئاخێنران/XN **po**:verb **is**:infinitive_intransitive_passive **st**:ئاخ | SFX N 0 [اە] کەی |
| ئاخێو/N **po**:noun **st**:ئاخ | SFX N 0 [اە] کەمان |

Table A1: Samples of lexicon entries to the right and morphological rules to the left in Hunspell for Sorani Kurdish. The `po`, `is` and `st` flags refer to part-of-speech tag, inflectional suffixes and stem, respectively. Morphophonological rules are provided in (Ahmadi, 2021).

## B SPARQL Queries

The following queries are used to extract Sorani Kurdish labels of cities (`Q515`), countries (`Q6256`) and humans (`Q5`) on Wikipedia using the Wikidata SPARQL endpoint. `ID` in the query is to be replaced by concept identifiers.

```
SELECT ?itemLabel
{
  ?item wdt:P31 wd:ID . # Q6256: country, Q5: human, Q515: city
  ?article schema:about ?item .
  ?article schema:isPartOf <https://ckb.wikipedia.org/> .
  SERVICE wikibase:label { bd:serviceParam wikibase:language "ckb" . }
   FILTER(EXISTS {
   ?item rdfs:label ?lang_label.
   FILTER(LANG(?lang_label) = "ckb")
   })
} LIMIT 100
```

## C Output of the Systems

The following table provides a few examples in the test sets ((Amani and Koochari, 2017; Mahmudi, 2019)) which are used for evaluating our baseline spell checker and our implementation in Hunspell. Various types of correct and incorrect words and suggestions are specified in different colors.

| Test set | Word | Correction | Suggestions: Our baseline | Suggestions: Hunspell |
|---|---|---|---|---|
| (Amani and Koochari, 2017) | بزوین | بزوێن | detected as correct | بووین – بوین – بزوێن – ببوین – بزوان – بروین – بوون – بکەین – بلێیین – بین |
| | سرماوەز | سەرماوەز | سەرماوەز – جەماوەر | سەرماوەز – سەرماوەزی – سەماوە – سەرماوە – سەماوەر – سورماوە – ماوەی – کراوە – لەماوەی |
| | مرج | مەرج | مەرج – مرد – مرۆ – فرج – مر – خرج – مرو – مرة – مرف – مری | مر – مەرج – مرت – مرن – مرم – مرک – مرگ – مرد – مرا – مرێ – مره – مری |
| | بوک | بووک | detected as correct | detected as correct |
| | وینە | وێنە | detected as correct | یونە – وەنی – نیوە – وتنە – وێنە – قینە – ونە – بوینە – وتینە – سوینە – وسینە – ورینە – وژینە – لۆینە – هوینە |
| | ووتن | وتن | detected as correct | توون – ووین – قوتن – وتن – وون – ووت – پووتن – تووتن – شووتن – سووتن – ڕووتن – فووتن – جووتن – لووتن – گووتن |
| | واژەدا | واژەدان | وانەدا – وادەدا – وێژەدا – کاتەدا – بوارەدا – وڵاتەدا – ڕۆژەدا – ساڵەدا – شارەدا – وڵاتەداواژەدان | detected as correct |
| | لەرزوک | لەرزۆک | لەرزۆک – لەرزێک – کەرکوک – لەرووی – لەنزیک – بەرزو – لەروی – هەرزوو – سەروک – وەرزێک | لەرزەک – لەرزک – لەرزۆک – لەرزێک – لەرزوو |
| (Mahmudi, 2019) | دیریکراوەکانی | دیاریکراوەکانی | داگیرکراوەکانی – هاوکاریکردنەکانی – داوەراندنەکانی | دیاریکراوەکانی – دیاریکراوەکان – دەرکراوەکانی – داگیرکراوەکانی – دیموکراتەکانی – دەرکراوەکان – دیاریکراوەکەی – داواکراوەکانی – دیاریکراوەکاندا – دیپۆرتکراوەکانی |
| | تەنها | تەنیا | detected as correct | detected as correct |
| | چاکبونەوەی | چاکبوونەوەی | detected as correct | چاکبوونەوەی – چاکبوونەوە – پاکبوونەوەی – چابوونەوەی – ناسکبوونەوەی |
| | دەربیکەن | دەری بکەن | detected as correct | detected as correct |
| | لەسەربکرێ | لەسەر بکرێت | detected as correct | لەسەر بکرێ – لە سەربکرێ – لەسەرخۆێ |
| | میلەتەکەمانە | میللەتەکەمانە | میلەتەکەمان – میللەتەکەمانە – میللەتەکەمان – میللەتەکەمانی – میلەتەکەیان – میللەتەکەمانن – میللەتەکەمانەو – میللەتەکەیان – میللەتەکەماندا | مەملەکەتانە |
| | پۆلیڤیاش | بۆلیڤیاش | پۆلیڤیا – پۆلیسیش – پۆلیسیان – بۆلیڤیا – ئۆلیڤیا – پۆلیسی – پۆلیسدا – پۆلۆنیا – لیبیاش – پۆلیسیی | ئۆلیڤیا |
| | یەکێتییەکەیانەوە | یەکێتییەکەیانەوە | یەکێتییەکانەوە – یەکێتییەکەیان – یەکێتیەکانەوە – یەکێتییەکەمان – یەکێتییەکەتان – یەکێتیەکەیان – یەکێتییەکانە – یەکێتییەکەوە – یەکێتییانەوە – یەکێتییەکان | detected as correct |
| | دەتواین | دەتوانین | دەتوانن – دەتوانین – دەتوان – دەدواین – دەتوانیت – دەبوایە – دەتوانم – دواین – نەتوانن – دەخوازن | دەتوانی – دەواین – دەتوان – دەتوانین – دەیتوان – دەسواین – دەڕواین – دەلواین – دەتناین – دەتوانن – دەتگاین – دەدواین – دەتداین – دەت واین – دە تواین |
| | ئاسانتتر | ئاسانتر | ئاسانتر – ئاسانترە – ئاساییتر – ئاسانکار – ئاسانتری – ئاسان – ئاسانی – ئاسانکاری – ئاسانە – ئەوانیتر | ئاسانترت – ئاسانتر – ئاسمانیتر |
| | سەپێندرابێت | سەپێنرابێت | سەپێنرابێت – سپێردرابێت – سەپێندرا – سەپێنرابوو – پێدرابێت – سەپێندراوی – بسەپێندرێت – سەپێندراوە – پێنەدرابێت – سەپاندبێت | detected as correct |

Legend
- Correct spelling
- Incorrect spelling
- Considered an incorrect spelling or prediction in the test set while is correct in some Sorani Kurdish sub-dialects
- Correct suggestion but not the expected correct replacement